%% file: iclr2025_conference.tex
\documentclass{article} 
\usepackage{iclr2025_conference,times}
\usepackage{float}
\input{math_commands.tex}

\usepackage{wasysym}
\usepackage{hyperref}
\usepackage{url}
\usepackage{booktabs}
\usepackage{graphicx}
\usepackage{float}
\usepackage{enumitem}
\usepackage{multirow}
\title{Think Big, Search Small: Where Capacity\\ Matters in Hierarchical Search Agents?}

\author{Qinnan Cai\thanks{Equal contribution.}, Yibo Zhao\footnotemark[1] \& Xiang Li \thanks{Corresponding author: \texttt{xiangli@dase.ecnu.edu.cn}.} \\
School of Data Science and Engineering\\
East China Normal University
}

\newcommand{\gain}[1]{\,{\scriptsize\color{green!50!black}(#1)}}

\definecolor{qonecolor}{RGB}{0,112,192}     
\definecolor{qtwocolor}{RGB}{192,0,0}       
\definecolor{qthreecolor}{RGB}{0,128,64}    
\newcommand{\qone}{\textcolor{qonecolor}{\textbf{Q1}}}
\newcommand{\qtwo}{\textcolor{qtwocolor}{\textbf{Q2}}}
\newcommand{\qthree}{\textcolor{qthreecolor}{\textbf{Q3}}}

\iclrfinalcopy 
\begin{document}

\maketitle

\begin{abstract}
Large language model based search agents increasingly adopt multi-agent architectures in which a main agent decomposes a complex question into sub-queries and dispatches them to parallel sub-agents. However, existing systems instantiate all roles from a single model of identical scale, leaving open how model capacity should be distributed across roles. We factorize hierarchical search into three roles: a \emph{delegation} role responsible for task decomposition, an \emph{execution} role responsible for retrieval and evidence extraction, and an \emph{answer generation} role held fixed as a confound control. We then conduct controlled capacity sweeps along the delegation and execution axes on five multi-hop QA benchmarks. The experiments yield three findings. First, role factorization consistently outperforms a single-agent baseline, improving exact match from 4.5 to 8.6 points across six model scales. Second, capacity sensitivity is asymmetric: scaling the delegation backbone improves EM by ${\sim}$11 points, whereas scaling the execution sub-agent moves EM by only ${\sim}$2.6 points, identifying decomposition as the capability bottleneck. Third, a 1.7B-parameter executor trained via quality-filtered trajectory distillation matches a frontier sub-agent in accuracy while consuming 37\% fewer sub-agent tokens, advancing the Pareto frontier. These results suggest a concrete recipe for building hierarchical search agents: concentrate capacity at delegation and downsize execution without sacrificing accuracy. Our code is available at \url{https://github.com/QinnanCai0115/role-factorized-search}.
\end{abstract}

\section{Introduction}
\label{sec:introduction}

Large language model (LLM) based search agents have rapidly emerged as a central topic in the LLM community~\citep{singh2025agentic,li2507towards,shi2025deep,team2026kimi}. By interleaving reasoning with retrieval, modern search agents can autonomously decompose a question, issue a sequence of queries, and synthesize evidence scattered across multiple retrieved documents~\citep{jin2025search,trivedi2023interleaving,yao2022react,li2025search,jiang2023active}. As both model capability and task ambition scale up, the scope of search keeps expanding from single-hop factoid questions~\citep{mallen2023not} to complex multi-hop questions whose answers must be composed from many pieces of evidence~\citep{yang2018hotpotqa,ho2020constructing,trivedi2022musique,press2023measuring}, making agentic search a demanding and informative testbed for LLM-based agents.

To date, the majority of search agents follow a single-agent route: one model is responsible for planning, issuing queries, reading retrieved documents, and producing the final answer, all within a single shared context~\citep{jin2025search,song2025r1,gao2025beyond}. This design is conceptually simple and has proven highly effective~\citep{singh2025agentic,li2507towards}. By construction, every retrieved passage and every intermediate observation accumulates in the same context window. 

However, as the number of hops and retrieval rounds grows, this context grows with it: passages that are each only locally relevant to one sub-question pile up alongside the original question and the agent's own plan~\citep{du2025context, liu2024lost}. The single model must therefore act as planner, reader, and synthesizer over a context that keeps expanding, a tension that grows more pronounced as questions require more reasoning hops and retrieval rounds.

A complementary route addresses this tension through multi-agent collaboration. Instead of concentrating all work in one context, these systems decompose a complex search task into sub-tasks and dispatch them to parallel sub-agents, each searching within an isolated context and returning only condensed findings~\citep{team2026kimi,xu2026wideseek,ning2026searchswarm}. Task decomposition thus doubles as context management: each sub-agent sees only the evidence relevant to its sub-task, while the main agent's context holds plans and summaries rather than raw passages. This route improves long-horizon and broad information-seeking tasks~\citep{xu2026wideseek,ning2026searchswarm}.

However, across existing multi-agent search systems, the question of how much model capacity each role actually requires has been left largely uninvestigated. These systems instantiate the main agent and the sub-agents from a single shared model of identical scale~\citep{xu2026wideseek,ning2026searchswarm}, and their effort is directed at making this one model a better delegator and executor, not at asking how capacity should be distributed between the two roles. This leaves a central question open: \emph{how should model capacity be allocated across roles in a hierarchical search agent to maximize the effectiveness--efficiency trade-off?} We decompose this into three sub-questions:

\begin{itemize}[leftmargin=*, itemsep=2pt, topsep=2pt, label={}]
\item \textbf{(\qone)~Does role factorization help?} Before studying allocation, we must verify in our controlled setting that separating delegation and execution is itself beneficial compared with a monolithic single-agent baseline.
\item \textbf{(\qtwo)~How to allocate capacity to reach the Pareto frontier?} Which role is the capability bottleneck, and how does the effectiveness--efficiency frontier shift as we redistribute capacity between delegation and execution?
\item \textbf{(\qthree)~How to advance the Pareto frontier?} Can targeted training of a compact executor push the frontier beyond what scaling alone achieves?
\end{itemize}

\noindent We argue that answering these questions is essential both for understanding the division of labor inside agentic systems and for building search agents that are effective and efficient at scale.

In this paper, we take a first step toward answering these questions. We explicitly factorize the search task into three roles: task delegation, where a main agent decomposes the user's question, plans, and dispatches focused sub-queries; search execution, where sub-agents carry out the actual querying, reading of retrieved documents, and evidence extraction; and answer synthesis, where a fixed module produces the final answer from the user question, the issued sub-queries, and their returned reports, without access to any agent's reasoning trajectory. The first two roles are the objects of our study; the third is held constant across all conditions, so that any change in performance reflects delegation or execution rather than answer writing. Built on this factorization, we conduct systematic experiments that vary the model capacity assigned to each role, and arrive at three principal findings:

\begin{itemize}[leftmargin=*, itemsep=0pt, topsep=0.2pt]
\item \textcolor{qonecolor}{\textbf{Multi-agent factorization consistently outperforms single-agent search.}}
Across six model scales, factoring the task into delegation and execution roles lifts exact match (EM) from 4.5 to 8.6 points over a single-agent baseline operating within one shared context (Sec.~\ref{sec:multi-vs-single}).

\item \textcolor{qtwocolor}{\textbf{Decomposition is the bottleneck; the Pareto frontier favors heterogeneous allocation.}}
Capacity sweeps along the two role axes reveal a stark asymmetry: scaling the backbone improves EM by ${\sim}$11 points, whereas scaling the sub-agent moves EM by only ${\sim}$2.6 points. The Pareto frontier concentrates capacity in the backbone while keeping the executor compact (Sec.~\ref{sec:execution-sweep}--\ref{sec:asymmetry-pareto}).

\item \textcolor{qthreecolor}{\textbf{A compact SFT executor advances the Pareto frontier.}}
Quality-filtered trajectory distillation trains a 1.7B-parameter executor that matches a frontier sub-agent while consuming fewer tokens, advancing the frontier beyond what scaling alone achieves (Sec.~\ref{sec:sft-executor}).
\end{itemize}

Together, these results suggest that the capability boundary of a hierarchical search agent is governed not by the uniform strength of its components, but by where capability is placed in the hierarchy. We hope this role-aware view provides empirical guidance for practitioners building multi-agent search systems and a conceptual lens for future research on agent abilities.

\section{Related Works}
\label{sec:related-works}

\subsection{Single-Agent Agentic Search}
\label{sec:single-agent}

Augmenting LLMs with retrieval has evolved from static retrieval-augmented generation into agentic search, where the model interleaves reasoning with iterative query issuing~\citep{yao2022react,trivedi2023interleaving,li2025search,jiang2023active}. Existing single-agent search systems fall into two architectural categories.
 
In most existing systems, a single model executes the entire search trajectory, from query formulation through retrieval to answer generation. Training approaches fall into three categories: reinforcement learning (RL) methods that optimize multi-turn search behavior directly against outcome rewards~\citep{chen2026learning,jin2025search,gao2025beyond}; supervised fine-tuning (SFT) methods that distill curated search trajectories into the model~\citep{sun2025simpledeepsearcher,du2026openseeker}; and hybrid SFT+RL pipelines that first warm-start the agent with supervised demonstrations before refining its search strategy via RL~\citep{song2025r1,wu2026webdancer,li2025websailorv2}. To isolate the search competence from answer writing, a second line of work decouples the search component from a frozen generator: a lightweight searcher is trained to retrieve and organize evidence, while answer generation is handled by a separate, fixed model~\citep{jiang2025s3,mei2025ai,jiang2025qagent,chen2025mao}. These modular designs validate that search can be isolated from answering, but the searcher itself remains a single sequential agent that must jointly handle planning, retrieval, and reading comprehension within one shared context.

Beyond single-context scaling, recent analyses reveal optimization and capacity conflicts when heterogeneous skills are placed inside one model~\citep{li2026reasoning,cai2026advancing}. DART~\citep{li2026reasoning} reports gradient interference between high-level reasoning and low-level tool calling, while modular gradient surgery~\citep{cai2026advancing} identifies analogous cross-task conflicts. These findings motivate decomposing the search process into separately instantiated roles, which we discuss next.

\subsection{Multi-Agent Search Architectures}
\label{sec:multi-agent-arch}

A natural response to these limitations is multi-agent collaboration: a lead agent decomposes the task and dispatches sub-tasks to parallel sub-agents, each operating in an isolated context and returning only condensed findings. WideSeek-R1~\citep{xu2026wideseek} trains such a lead-sub framework with multi-agent RL for broad information seeking, jointly optimizing the lead agent and parallel sub-agents that share one LLM with isolated contexts. SearchSwarm~\citep{ning2026searchswarm} synthesizes harness-guided delegation trajectories and internalizes ``delegation intelligence'' into model weights via SFT for long-horizon deep research; industrial systems report similar orchestrator-worker designs~\citep{team2026kimi}. However, these systems instantiate the orchestrator and the executors with models of the same scale, typically a single shared LLM, leaving open which role actually demands capacity and whether the executor can be radically downsized.

Evidence from adjacent domains suggests that such downsizing may be viable. Surveys of small language models in agentic settings argue that compact models can be effective for narrow, well-specified functions such as tool invocation and structured extraction, while larger models remain more useful for open-ended planning and reasoning~\citep{sharma2025small}. This view aligns with the T2 paradigm of agent-supervised tool adaptation, which keeps a core agent frozen while adapting external tools or sub-agents using signals from that agent~\citep{jiang2025adaptation}. A concrete example is Terminus-4B~\citep{garg2026terminus}, which trains a 4B terminal-execution sub-agent for coding agents and matches frontier-model sub-agents at lower cost. These results suggest that execution-like components can often be specialized and compressed, but they are established outside hierarchical search and do not determine how capacity should be allocated across the delegation--execution hierarchy of a search agent. We provide such a characterization by varying the capacity assigned to each role independently and showing that decomposition quality, rather than executor scale, governs the capability boundary in our controlled setting.

\section{Controlled Study Design}
\label{sec:study-design}
 
We adopt a role-factorized search design as an experimental instrument for isolating how model capacity affects delegation and execution in hierarchical search. This section describes the instrument (Sec.~\ref{sec:role-factorization}), defines the experimental variables and controlled factors (Sec.~\ref{sec:variables}), specifies the intervention protocol that structures our analysis (Sec.~\ref{sec:intervention-protocol}), and presents a training pipeline for a compact executor that tests whether the implied allocation is realizable in practice (Sec.~\ref{sec:training-compact-executors}).
 
\begin{figure}[t]
    \centering
    \includegraphics[width=\linewidth]{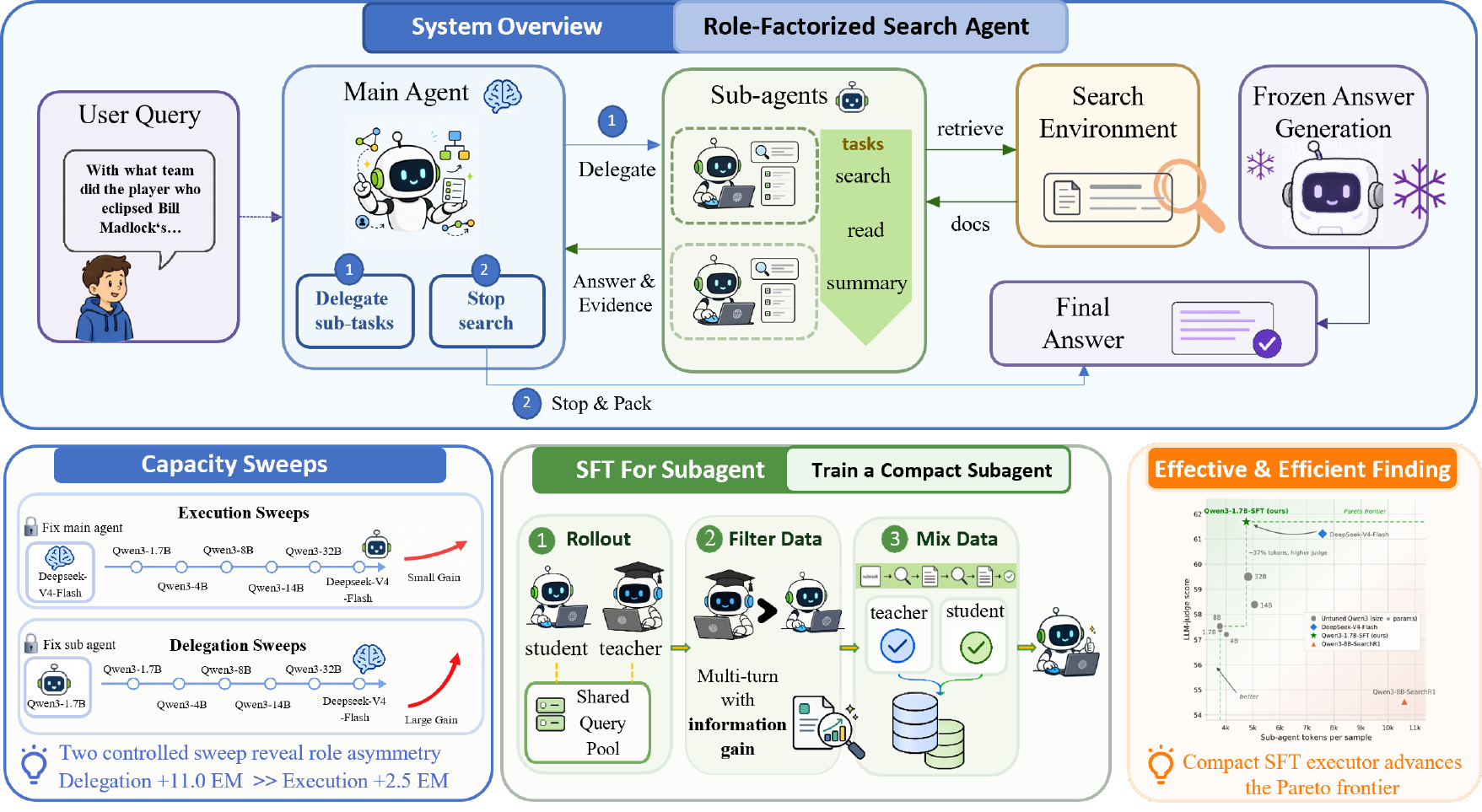}
    \caption{Overview of our hierarchical search framework. The delegation policy $\pi_D$ (backbone) decomposes the question into sub-queries; the execution policy $\pi_E$ (sub-agent) retrieves and reads within an isolated context, returning only a compressed report; a fixed $\pi_A$ synthesizes the final answer. We treat $(c_D, c_E)$ as the independent variable, holding $\pi_A$ constant.}
    \label{fig:method}
\end{figure}
 
\subsection{Role-Factorized Search as an Experimental Instrument}
\label{sec:role-factorization}

We model hierarchical search as three role-specific policies (Fig.~\ref{fig:method}): a \textbf{delegation} policy $\pi_D$ realized by the main agent, an \textbf{execution} policy $\pi_E$ realized by the search sub-agent, and an \textbf{answer-generation} policy $\pi_A$ realized by a fixed answerer; the subscripts $D$, $E$, and $A$ abbreviate the three role names. Given a user question $q$, the system produces an evidence-grounded answer $y$ through the following role-separated interface.

\paragraph{Delegation (main agent).}
The main agent follows $\pi_D$ and maintains its own context $H^D$. At each round, up to a budget of $T$ delegation rounds, it either dispatches sub-queries or terminates the search. A \textbf{sub-query} $q_i$ is a self-contained atomic sub-question obtained by decomposing $q$ (typically one attribute or relation of one entity), interpretable without access to $H^D$; the full decomposition rules are given in App.~\ref{app:role-prompts}. The main agent decomposes $q$, integrates the returned reports into its plan, and decides when enough evidence has been gathered; it does not write the final answer.

\paragraph{Execution (sub-agent).}
Each sub-query $q_i$ is handled by an independent sub-agent following $\pi_E$, instantiated in a fresh context $H^E$ that contains only $q_i$. The sub-agent runs an inner ReAct-style search loop~\citep{yao2022react} for at most $K$ assistant turns; its prompt (App.~\ref{app:role-prompts}) requires the answer to be grounded in the retrieved documents rather than parametric memory. On termination, it emits a \textbf{report} $r_i$, a concise answer to $q_i$ with supporting evidence snippets; $r_i$ is the only information returned to the main agent.

\paragraph{Answer generation and context isolation.}
When the main agent stops after $T' \leq T$ delegation rounds, the question $q$ and the collected sub-query--report pairs $R=\{(q_i, r_i)\}_{i=1}^{N}$, where $N$ is the total number of dispatched sub-queries, are passed with an answer-generation instruction to $\pi_A$, which produces the final answer $y$. We hold $\pi_A$ fixed across all conditions; its instantiation is given in Sec.~\ref{sec:experimental-setup}. The answerer sees neither the main agent's reasoning nor the sub-agents' trajectories or raw retrieved passages. This isolation is a deliberate confound control against reasoning leakage into answer writing: with $\pi_A$ fixed and its input restricted to $q$ and $R$, variation along the delegation axis reflects decomposition quality, while variation along the execution axis reflects the quality of retrieval and evidence extraction.
\subsection{Capacity Variables and Controlled Factors}
\label{sec:variables}
 
To study how model capacity should be distributed between the delegation and execution roles, we write $c_D$ and $c_E$ for the capacity (scale) of the model assigned to each role, holding the answerer capacity $c_A$ fixed throughout. Whereas existing multi-agent search systems implicitly fix $c_D = c_E$ by drawing both roles from one shared model (Sec.~\ref{sec:multi-agent-arch}), we treat the pair $(c_D, c_E)$ as the sole \textbf{independent variable} of our study, and ask how end-task accuracy, inference cost, and the \emph{location of the capability bottleneck} change as a function of this allocation. All other factors, i.e., the answer-generation model, the retriever and corpus, prompts, search budgets, the decoding protocol, and the evaluation suite, are held fixed across all conditions; their concrete instantiation is given in Sec.~\ref{sec:experimental-setup}. This design ensures that any observed performance difference reflects the capacity of the delegation or execution role, rather than differences in answer writing, retrieval quality, or evaluation noise.
 
\subsection{Intervention Protocol}
\label{sec:intervention-protocol}
 
Rather than densely populating the full $(c_D, c_E)$ grid, we define three targeted interventions, each designed to answer one research question while holding all other factors constant.
 
\paragraph{Intervention 1: Role factorization.}
Does separating delegation and execution improve over a single shared-context agent? We compare (i) a single-agent baseline that performs decomposition, retrieval, and answer generation within one context window, against (ii) our role-factorized setup with the same model instantiated in both the delegation and execution roles. This intervention further examines \qone{} in our controlled setting and establishes whether the factorization itself carries value independent of capacity allocation.
 
\paragraph{Intervention 2: Delegation-capacity sweep.}
How sensitive is end-task performance to backbone capacity? We fix the executor at Qwen3-1.7B and the answerer at Qwen3-32B, then vary the backbone across models of increasing scale. Since the executor and $\pi_A$ are both held constant, any change in performance is attributable to the backbone's decomposition quality alone.
 
\paragraph{Intervention 3: Execution-capacity sweep.}
How sensitive is performance to executor capacity? We fix the backbone at DeepSeek-V4-Flash and the answerer at Qwen3-32B, then vary the executor across models of increasing scale, including our SFT-trained compact executor (Sec.~\ref{sec:training-compact-executors}). Symmetrically, any performance change is attributable to execution quality alone.
 
\paragraph{Interpreting the contrast.}
Interventions 2 and 3 each vary one role while holding the other two fixed; contrasting their accuracy responses therefore localizes the capability bottleneck, and the contrast cannot be confounded by differences in answer writing because $\pi_A$ is constant throughout. We present the contrast in Sec.~\ref{sec:asymmetry-pareto}.
 
\subsection{Training a Compact Executor}
\label{sec:training-compact-executors}

The interventions above reveal \emph{where} the capability bottleneck lies. Even if execution proves far less capacity-sensitive than delegation \qtwo, an untuned compact executor still trails a frontier one, and the two roles call for different remedies. Delegation demands broad world knowledge and general long-horizon reasoning, so the backbone is best kept as a strong general model; fine-tuning it for search risks degrading these general capabilities. Execution, by contrast, is a narrow and well-scoped skill that a compact specialized model could plausibly master. This motivates \qthree: we ask \emph{where a compact executor falls short of a frontier one, and how targeted training can compensate for this gap}.

Within our framework, a sub-agent operates under well-scoped sub-queries issued by the backbone and faces two categories of execution. In \textbf{single-search} cases the sub-query is straightforward enough that one retrieval round suffices: the agent formulates a search query, reads the retrieved passages, and returns a report. In \textbf{multi-search} cases the initial retrieval is insufficient (the top passages lack the answer, the query needs refinement, or additional evidence is required) and the agent must recognize the gap, reformulate its query, and issue a second retrieval round before reporting. Small models generally handle single-search competently: given a well-scoped sub-query, even a 1.7B model can formulate an adequate retrieval query and extract the answer from the returned passages. The critical gap lies in multi-search correction, i.e., recognizing that the first attempt was insufficient and constructing an effective follow-up query, a capability that demands reasoning about what information is still missing.

This gap analysis motivates the training design: preserve the student's existing single-search competence while selectively injecting the multi-search correction behavior it lacks, using targeted trajectory distillation from a stronger teacher. We instantiate the student as Qwen3-1.7B~\citep{yang2025qwen3} and collect supervision by rolling out a teacher model, DeepSeek-V4-Pro~\citep{deepseekai2026deepseekv4}, which is stronger than the DeepSeek-V4-Flash sub-agent evaluated in Sec.~\ref{sec:analysis}, placed at the sub-agent position under a DeepSeek-V4-Flash backbone.

\paragraph{Trajectory rollout.}
For each training query, the fixed backbone (DeepSeek-V4-Flash) generates a sub-query and dispatches it to both the teacher (DeepSeek-V4-Pro) and the untuned student (Qwen3-1.7B), each operating under the execution prompt of App.~\ref{app:role-prompts}. We record both models' full execution trajectories (reasoning traces, retrieval actions, retrieved passages, and the final report) for use in the quality filtering stage below.

\paragraph{Quality filtering.}
Guided by the capability gap identified above, we construct a targeted SFT corpus from two trajectory strata with two design principles: preserve the student's existing single-search competence, and selectively inject the multi-search correction behavior it lacks. Throughout, a trajectory is labeled \emph{correct} only if the final answer it yields attains EM${}=1$ and is accepted by the LLM judge (App.~\ref{app:judge-prompt}).

\begin{itemize}[leftmargin=*, itemsep=2pt, topsep=2pt]
\item \textbf{Correct single-search.} Single-retrieval trajectories from either the teacher or the untuned student that satisfy the correctness criterion (EM${=}1$ and accepted by the LLM judge). Retaining the student's own correct trajectories anchors its existing capability and guards against regression during fine-tuning; the teacher's correct trajectories supplement coverage on cases the student does not already handle.
\item \textbf{Teacher multi-search with information gain.} Two-search trajectories for which a single retrieval round is insufficient: we re-run the teacher with its search budget capped at one round and retain only cases where this single-search variant fails to produce a correct answer while the full two-search trajectory succeeds. This counterfactual filter ensures that every included multi-search demonstration reflects a genuine need for iterative retrieval, avoiding redundant second-round searches that would teach the student to retry unnecessarily.
\end{itemize}

\paragraph{Training objective.}
We fine-tune Qwen3-1.7B on the filtered corpus with standard SFT, computing the loss only on model-generated tokens (reasoning traces, retrieval actions, and the final report) and masking out the retrieved passages. This masking ensures that the model learns to reason about and act on retrieved evidence rather than to memorize it. The composition of the SFT corpus is reported in App.~\ref{app:sft-data}.

\section{Analysis}
\label{sec:analysis}
 
We organize the analysis around the three questions posed in Sec.~\ref{sec:introduction}, holding the answer-generation module fixed and varying only the model placed at the delegation and execution roles. We first establish that role factorization itself improves search (Sec.~\ref{sec:multi-vs-single}, \qone). We then conduct systematic capacity sweeps along the delegation axis (Sec.~\ref{sec:delegation-sweep}) and execution axis (Sec.~\ref{sec:execution-sweep}), and compare the two to characterize the Pareto frontier of capacity allocation (Sec.~\ref{sec:asymmetry-pareto}, \qtwo). Finally, we show that a compact SFT-trained executor advances this frontier, matching frontier execution at a fraction of the cost (Sec.~\ref{sec:sft-executor}, \qthree), and provide qualitative evidence through a case study (Sec.~\ref{sec:case-study}).
 
\subsection{Experimental Setup}
\label{sec:experimental-setup}
 
\noindent\textbf{Benchmarks and metrics.}
We adopt the evaluation suite and retrieval environment of~\citet{zhao2026retrieval}: we evaluate on 2WikiMultihopQA~\citep{ho2020constructing}, HotpotQA~\citep{yang2018hotpotqa}, MuSiQue~\citep{trivedi2022musique} (up to 1,000 instances each), PopQA~\citep{mallen2023not} (1,000 instances), and Bamboogle~\citep{press2023measuring} (125 instances), following the same five-benchmark protocol. From these 4{,}125 instances we exclude 256 that the delegation agent answers directly without issuing any sub-query, since such instances are solvable from parametric memory alone; all experiments use the remaining 3{,}869 instances (per-dataset breakdown in App.~\ref{app:eval-cleaning}). We report Exact Match (EM), token-level F1, and an LLM-as-judge score following~\citet{song2025r1}, which uses DeepSeek-V4-Flash to assess whether the predicted answer aligns with the meaning and key information of the gold answer, crediting semantically correct answers missed by string matching (prompt in App.~\ref{app:judge-prompt}). For each configuration, we sample four independent rollouts per query and report the mean (mean@4) to reduce variance from stochastic decoding. Unless otherwise noted, all reported numbers are averaged across the five benchmarks.
 
\noindent\textbf{System instantiation.}
We use the minimal main--sub architecture of Sec.~\ref{sec:role-factorization}. Retrieval is served by a Qwen3-Embedding-8B~\citep{yang2025qwen3} dense retriever over the \textbf{Wiki-fixed} corpus of~\citet{zhao2026retrieval}, a completed version of the Wikipedia-2018 dump that restores supporting documents missing from Wiki-18, indexed and served through FAISS~\citep{douze2024faiss}. We cap the outer delegation loop at $T{=}4$ rounds and each inner execution loop at $K{=}3$ assistant turns. The answer-generation module $\pi_A$ is instantiated as Qwen3-32B~\citep{yang2025qwen3} and held fixed across every configuration. We choose Qwen3-32B over frontier alternatives for two reasons: (i)~frontier models such as DeepSeek may exhibit data contamination with the evaluation benchmarks, which would blur the distinction between ``the agent retrieved good evidence'' and ``the model already knew the answer from pretraining''; (ii)~Qwen3-32B synthesizes the collected reports faithfully without discarding useful evidence, while its no-retrieval accuracy remains low enough to confirm that it does not rely on parametric memory (verified by a no-retrieval control on the test set in App.~\ref{app:answerer-choice}).
 
\noindent\textbf{Reference configurations.}
Beyond varying model scale at each role, we evaluate three reference configurations: (i) Direct Tool, in which the main agent calls the retriever itself with no sub-agent (the single-agent setting of Tab.~\ref{tab:single_to_mainsub_gain}); (ii) a frontier DeepSeek-V4-Flash sub-agent, representing an execution role with no capacity constraint; and (iii) 
 Qwen3-8B-SearchR1~\citep{zhao2026retrieval}, a search agent trained end-to-end on multi-hop QA via reinforcement learning, placed at the execution role to investigate whether an agent trained outside the delegation--execution framework is qualified as an executor, i.e., whether such off-the-shelf searchers can substitute for the in-framework training of Sec. 3.4.
\subsection{Role Factorization Improves Search}
\label{sec:multi-vs-single}

Before dissecting the two roles, we first ask whether the factorization itself adds value (Intervention~1, Sec.~\ref{sec:intervention-protocol}): the same model either performs the entire task as a single agent within one shared context, or serves as both backbone and sub-agent in our main--sub architecture. We run this comparison on six models spanning Qwen3-1.7B to DeepSeek-V4-Flash and report results in Tab.~\ref{tab:single_to_mainsub_gain}.
 
\begin{table}[t]
\centering
\small
\setlength{\tabcolsep}{5.5pt}
\renewcommand{\arraystretch}{1.10}
\caption{Performance gains from single-agent to main--sub architecture (3,869-instance test set).}
\label{tab:single_to_mainsub_gain}
\begin{tabular}{lccc ccc}
\toprule
\multirow{2}{*}[-0.4ex]{\textbf{Model}}& \multicolumn{3}{c}{\textbf{Single Agent}}
& \multicolumn{3}{c}{\textbf{Main--Sub Agent}} \\
\cmidrule(lr){2-4}
\cmidrule(lr){5-7}
& EM & F1 & Judge
& EM & F1 & Judge \\
\midrule
Qwen3-1.7B
& 20.24 & 28.38 & 32.64
& \textbf{27.86}\gain{+7.62}
& \textbf{37.54}\gain{+9.16}
& \textbf{41.50}\gain{+8.86} \\
Qwen3-4B
& 21.85 & 29.65 & 32.04
& \textbf{27.32}\gain{+5.47}
& \textbf{36.12}\gain{+6.47}
& \textbf{40.41}\gain{+8.36} \\
Qwen3-8B
& 24.32 & 31.91 & 35.41
& \textbf{31.71}\gain{+7.39}
& \textbf{41.31}\gain{+9.41}
& \textbf{45.28}\gain{+9.87} \\
Qwen3-14B
& 27.92 & 37.51 & 41.01
& \textbf{36.55}\gain{+8.63}
& \textbf{46.37}\gain{+8.86}
& \textbf{51.91}\gain{+8.91} \\
Qwen3-32B
& 31.38 & 42.78 & 50.71
& \textbf{36.86}\gain{+5.48}
& \textbf{47.80}\gain{+5.02}
& \textbf{54.65}\gain{+3.94} \\
DeepSeek-V4-Flash
& 37.24 & 49.06 & 57.89
& \textbf{41.77}\gain{+4.52}
& \textbf{53.34}\gain{+4.28}
& \textbf{61.21}\gain{+3.32} \\
\bottomrule
\end{tabular}
\vspace{-0.4em}
\end{table}
 
The main--sub factorization outperforms the single-agent baseline at every scale, with EM gains ranging from $+4.52$ (DeepSeek-V4-Flash) to $+8.63$ (Qwen3-14B) and consistent improvements in F1 and LLM-judge. Two observations are worth noting. First, every Qwen3 model gains at least $5.4$ EM points, whereas the frontier model gains the least, suggesting that role factorization is particularly beneficial when a single model lacks the capacity to jointly handle planning, retrieval, and synthesis within one context. Second, even the frontier model benefits, indicating that the architectural separation of delegation from execution provides value beyond what raw model scale can deliver. These results establish that multi-agent role factorization is a meaningful architectural choice, not merely an artifact of using stronger models, and motivate the finer-grained capacity analysis in subsequent sections.
 
\subsection{Delegation Capacity Sets the Performance Ceiling}
\label{sec:delegation-sweep}
 
We fix the execution role to the naive Qwen3-1.7B and scale the backbone (Intervention~2). Tab.~\ref{tab:delegation-sweep} shows that delegation capacity matters a great deal.
 
\begin{table*}[t]
\centering
\caption{Delegation capacity sweep with Qwen3-1.7B as sub-agent (mean@4). Best in \textbf{bold}, second best \underline{underlined}.}
\label{tab:delegation-sweep}
\small
\setlength{\tabcolsep}{5pt}
\begin{tabular}{@{}l ccccc ccc@{}}
\toprule
\multirow{2}{*}[-0.45ex]{\textbf{Backbone}}& \multicolumn{5}{c}{LLM Judge by Dataset} & \multicolumn{3}{c}{Avg.} \\
\cmidrule(lr){2-6} \cmidrule(lr){7-9}
 & 2Wiki & HotpotQA & MuSiQue & PopQA & Bamb. & EM & F1 & Judge \\
\midrule
Qwen3-1.7B        & 55.57 & 52.55 & 18.24 & \underline{39.85} & 39.45 & 27.86 & 37.54 & 41.50 \\
Qwen3-4B           & 57.21 & 46.94 & 19.03 & 36.43 & 38.53 & 28.04 & 36.67 & 39.98 \\
Qwen3-8B           & 63.87 & 51.82 & 20.87 & 36.43 & 41.28 & 29.59 & 39.24 & 43.46 \\
Qwen3-14B          & 66.06 & 58.57 & \underline{26.92} & 39.12 & 44.04 & 32.26 & 42.51 & 47.88 \\
Qwen3-32B          & \underline{69.73} & \underline{60.71} & 26.52 & 38.39 & \underline{49.54} & \underline{33.94} & \underline{44.45} & \underline{49.25} \\
DeepSeek-V4-Flash  & \textbf{75.38} & \textbf{70.89} & \textbf{35.36} & \textbf{46.70} & \textbf{54.13} & \textbf{39.18} & \textbf{50.72} & \textbf{57.38} \\
\bottomrule
\end{tabular}
\end{table*}
 
With the executor and answerer fixed, Avg.\ EM rises \emph{monotonically} with delegation capacity: every step up the backbone scale improves EM, from $27.86$ with a Qwen3-1.7B backbone to $39.18$ with DeepSeek-V4-Flash, a gain of more than eleven points. LLM-judge follows the same overall trend, climbing by nearly sixteen points (from $41.50$ to $57.38$), with the same 1.7B executor doing all of the searching. The quality of decomposition, i.e., how the backbone decomposes the question into sub-queries, is the sole variable that changes, and it changes outcomes dramatically. This steep and consistent response confirms that decomposition quality sets the performance ceiling of the hierarchy, even when the executor is held constant at the smallest scale.
 
\subsection{Execution Capacity Has Diminishing Returns}
\label{sec:execution-sweep}
 
We now turn to Intervention~3: fix the backbone to DeepSeek-V4-Flash and vary only the execution role. Tab.~\ref{tab:execution-sweep} reports per-dataset LLM-judge scores and averaged metrics.
 
\begin{table*}[t]
\centering
\caption{Execution capacity sweep with DeepSeek-V4-Flash as backbone (mean@4). Best in \textbf{bold}, second best \underline{underlined}.}
\label{tab:execution-sweep}
\small
\setlength{\tabcolsep}{5pt}
\begin{tabular}{@{}l ccccc ccc@{}}
\toprule
\multirow{2}{*}[-0.45ex]{\textbf{Sub-agent}}& \multicolumn{5}{c}{LLM Judge by Dataset} & \multicolumn{3}{c}{Avg.} \\
\cmidrule(lr){2-6} \cmidrule(lr){7-9}
 & 2Wiki & HotpotQA & MuSiQue & PopQA & Bamb. & EM & F1 & Judge \\
\midrule
Qwen3-1.7B        & 75.38 & 70.89 & 35.36 & \underline{46.70} & 54.13 & 39.18 & 50.72 & 57.38 \\
Qwen3-4B           & 74.17 & 72.42 & 35.70 & 45.72 & 49.54 & 40.01 & 51.03 & 57.21 \\
Qwen3-8B           & 74.44 & 70.79 & 37.35 & 45.60 & 59.63 & 41.04 & 51.87 & 57.54 \\
Qwen3-14B          & 76.06 & 72.35 & 37.65 & 45.60 & 58.72 & 40.68 & 51.88 & 58.39 \\
Qwen3-32B          & 77.60 & 72.56 & 39.23 & \textbf{46.94} & 56.88 & 40.45 & 52.12 & 59.50 \\
DeepSeek-V4-Flash  & \underline{80.73} & \underline{75.29} & \textbf{41.30} & 44.74 & \underline{63.30} & \underline{41.77} & \underline{53.34} & \underline{61.21} \\
\midrule
Qwen3-8B-SearchR1  & 72.09 & 68.74 & 33.23 & 42.18 & 55.96 & 41.51 & 50.89 & 54.55 \\
Qwen3-1.7B-SFT    & \textbf{81.13} & \textbf{75.60} & \underline{41.09} & 46.21 & \textbf{64.80} & \textbf{41.81} & \textbf{53.97} & \textbf{61.69} \\
\bottomrule
\end{tabular}
\vspace{2mm}
\end{table*}
 
Two observations stand out. First, execution capacity shows diminishing returns among untuned models. Scaling from Qwen3-1.7B to Qwen3-32B moves Avg.\ EM by only $1.27$ points (from $39.18$ to $40.45$) and LLM-judge by about two points (from $57.38$ to $59.50$); even the frontier DeepSeek-V4-Flash sub-agent lifts EM by only $2.59$ total. This pattern is consistent across all five benchmarks. Once the backbone produces well-scoped sub-queries, execution is necessary but far less capacity-sensitive: a compact executor solves them almost as well as a much larger one.

Second, end-to-end search training does not transfer to the executor role. Qwen3-8B-SearchR1 achieves competitive EM ($41.51$) but the lowest LLM-judge score ($54.55$), below even the untuned Qwen3-1.7B ($57.38$). We attribute this to a two-fold mismatch between its training and the executor role. Logically, Search-R1 is trained to plan, decompose, and answer a full question on its own, so when handed an already-scoped sub-query it tends to re-plan and over-search rather than execute; distributionally, it is trained on complete multi-hop questions, whereas the executor receives atomic sub-queries drawn from a different input distribution. This mismatch is precisely why we roll out and filter trajectories within the delegation--execution framework (Sec.~\ref{sec:training-compact-executors}) instead of reusing an off-the-shelf searcher: the in-framework SFT-trained Qwen3-1.7B-SFT achieves the highest EM ($41.81$) and judge ($61.69$) among all sub-agents, surpassing even the frontier DeepSeek-V4-Flash, foreshadowing the Pareto frontier analyzed in the next section.
 
\subsection{Role Asymmetry and the Pareto Frontier}
\label{sec:asymmetry-pareto}

\begin{figure}[t]
\centering
\includegraphics[width=\columnwidth]{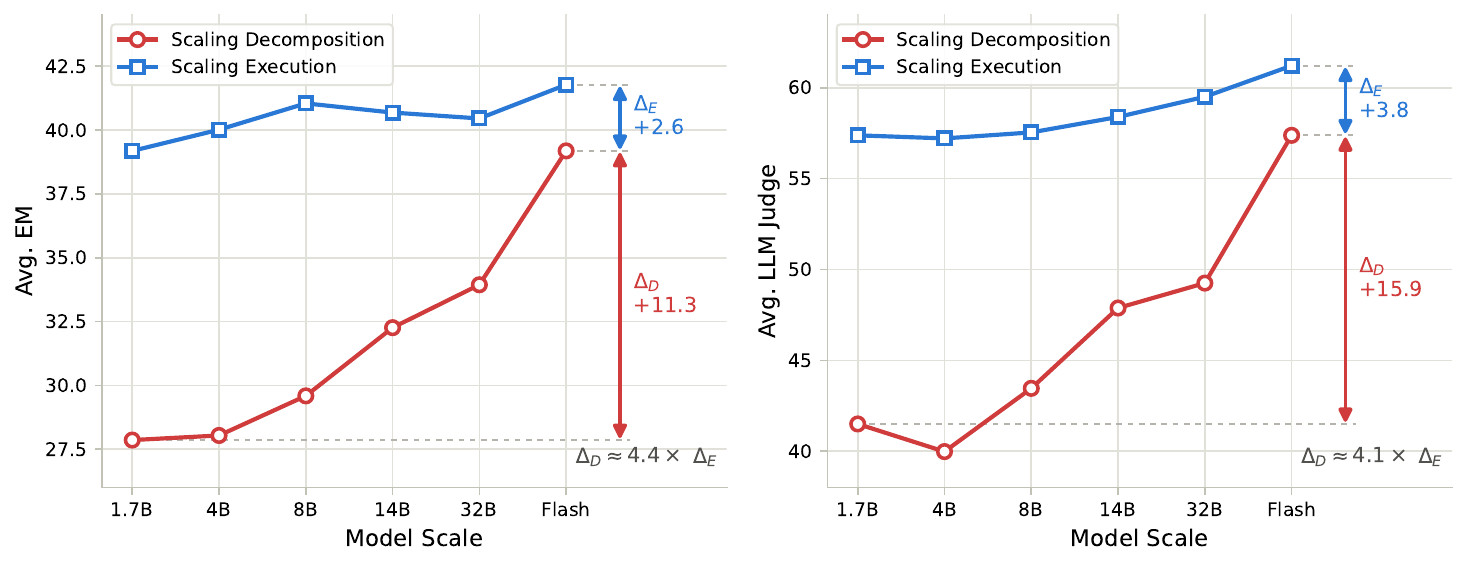}
\caption{Capacity sensitivity of each role. \textit{Scaling Execution} ({\color{blue}blue}) fixes the backbone to DeepSeek-V4-Flash and varies the sub-agent; \textit{Scaling Decomposition} ({\color{red}red}) fixes the sub-agent to Qwen3-1.7B and varies the backbone. Arrows mark each sweep's total gain ($\Delta_D$ vs.\ $\Delta_E$).}
\label{fig:asymmetry}
\end{figure}

Comparing the two sweeps reveals a stark asymmetry, visualized in Fig.~\ref{fig:asymmetry}. The execution curve (blue) is nearly flat: holding decomposition fixed and scaling the sub-agent from 1.7B to frontier scale moves EM by only about two and a half points ($\Delta$EM\,$\approx 2.6$, $\Delta$Judge\,$\approx 3.8$; Sec.~\ref{sec:execution-sweep}). The decomposition curve (red) rises steeply: holding execution fixed and scaling the backbone over the same range moves EM by more than eleven points (Sec.~\ref{sec:delegation-sweep}), a more than four-fold difference in sensitivity that is consistent across F1 and LLM-judge. At the smallest backbone, the gap between the two curves exceeds eleven EM points and fifteen judge points; it closes only when the backbone itself reaches frontier scale. Because $\pi_A$ is fixed throughout, neither effect can be ascribed to answer writing: the capability boundary of a hierarchical search agent is set by decomposition quality, not execution capability.
 
\begin{figure}[t]
\centering
\includegraphics[width=0.8\columnwidth]{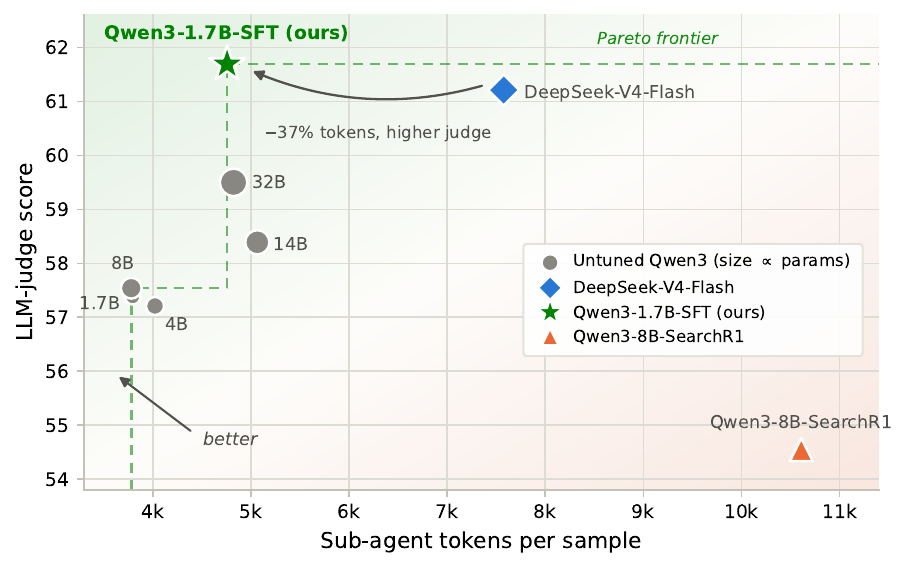}
\caption{Effectiveness--efficiency trade-off across sub-agent configurations under a DeepSeek-V4-Flash backbone. SFT-Qwen3-1.7B exceeds the frontier sub-agent in judge score while consuming $37\%$ fewer tokens with a model two orders of magnitude smaller.}
\label{fig:pareto}
\end{figure}
 
This asymmetry directly shapes the Pareto frontier of capacity allocation: because execution is insensitive to scale, the executor can be aggressively downsized without sacrificing accuracy, while capacity is concentrated in the backbone where it drives gains. Fig.~\ref{fig:pareto} visualizes this trade-off. Among all sub-agent configurations, SFT-Qwen3-1.7B achieves a judge score exceeding the frontier DeepSeek-V4-Flash sub-agent ($61.69$ vs.\ $61.21$) while consuming $37\%$ fewer sub-agent tokens ($4{,}758$ vs.\ $7{,}577$) and using a model with ${\sim}167\times$ fewer parameters. In contrast, the RL-trained Search-R1 consumes $2.2\times$ more tokens than SFT-Qwen3-1.7B yet scores the lowest on LLM-judge ($54.55$). The Pareto frontier thus favors a strong backbone paired with a compact, task-specialized executor, a finding we validate further in the next section.
 
\subsection{Compact Executor Training Realizes the Allocation}
\label{sec:sft-executor}

The preceding analysis establishes that decomposition is the bottleneck and that the executor can be radically downsized. Before evaluating the SFT executor, we first verify the capability gap hypothesized in Sec.~\ref{sec:training-compact-executors} by examining the internal search behavior of the untuned student (Qwen3-1.7B) and the teacher (DeepSeek-V4-Pro) during trajectory rollout on the training queries, both operating under the same DeepSeek-V4-Flash backbone.

\begin{table}[t]
\centering
\caption{Internal search behavior of the student and teacher sub-agents during trajectory rollout on 3{,}000 training queries (backbone: DeepSeek-V4-Flash).}
\label{tab:search-behavior}
\small
\setlength{\tabcolsep}{4pt}
\begin{tabular}{@{}l cc cc cc@{}}
\toprule
\multirow{2}{*}[-0.45ex]{\textbf{Sub-agent}}& \multirow{2}{*}[-0.45ex]{Invocations} & \multirow{2}{*}[-0.45ex]{\shortstack[c]{Avg. searches\\ per invocation}}  & \multicolumn{2}{c}{1-search} & \multicolumn{2}{c}{$\geq$2-search} \\
\cmidrule(lr){4-5} \cmidrule(lr){6-7}
 &  & & $n$ & \% & $n$ & \% \\
\midrule
Qwen3-1.7B       & 6{,}348 & 1.002 & 6{,}337 & 99.83 & 11 & 0.17 \\
DeepSeek-V4-Pro   & 6{,}189 & 1.675 & 3{,}694 & 59.69 & 2{,}495 & 40.31 \\
\bottomrule
\end{tabular}
\end{table}

Tab.~\ref{tab:search-behavior} reveals a stark behavioral difference. The naive Qwen3-1.7B performs single-search in 99.83\% of invocations, issuing a second retrieval round in only 11 out of 6{,}348 cases. DeepSeek-V4-Pro, by contrast, uses two or more searches in 40.31\% of invocations, with an average of 1.675 searches per invocation. This confirms the capability gap identified in Sec.~\ref{sec:training-compact-executors}: the small model lacks the multi-search correction ability, i.e.\ recognizing that the first retrieval was insufficient and reformulating a follow-up query, that accounts for a substantial portion of the performance difference between the two scales. The SFT training pipeline is designed precisely to inject this multi-search behavior while preserving the student's existing single-search competence.

We now evaluate whether this training succeeds. We pair three executors (naive Qwen3-1.7B, Qwen3-1.7B-SFT, and DeepSeek-V4-Flash) with two backbones and report results in Tab.~\ref{tab:sft-effectiveness}.

\begin{table*}[t]
\centering
\small
\setlength{\tabcolsep}{5pt}
\caption{SFT executor effectiveness across two backbones (mean@4). Qwen3-1.7B-SFT matches or exceeds the frontier sub-agent under both.}\label{tab:sft-effectiveness}
\begin{tabular}{@{}l ccccc ccc@{}}
\toprule
\multirow{2}{*}[-0.45ex]{\textbf{Sub-agent}}& \multicolumn{5}{c}{LLM Judge by Dataset} & \multicolumn{3}{c}{Avg.} \\
\cmidrule(lr){2-6} \cmidrule(lr){7-9}
 & 2Wiki & HotpotQA & MuSiQue & PopQA & Bamb. & EM & F1 & Judge \\
\midrule
\multicolumn{9}{@{}l}{\textit{Backbone: DeepSeek-V4-Flash}} \\[2pt]
\quad Qwen3-1.7B          & 75.38 & 70.89 & 35.36 & \textbf{46.70} & 54.13 & 39.18 & 50.72 & 57.38 \\
\quad Qwen3-1.7B-SFT      & \textbf{81.13} & \textbf{75.60} & 41.09 & 46.21 & \textbf{64.80} & \textbf{41.81} & \textbf{53.97} & \textbf{61.69} \\
\quad DeepSeek-V4-Flash    & 80.73 & 75.29 & \textbf{41.30} & 44.74 & 63.30 & 41.77 & 53.34 & 61.21 \\
\midrule
\multicolumn{9}{@{}l}{\textit{Backbone: GLM-5.1}} \\[2pt]
\quad Qwen3-1.7B          & 75.96 & 72.27 & 38.20 & 45.97 & 56.88 & 38.92 & 51.20 & 58.52 \\
\quad Qwen3-1.7B-SFT      & \textbf{81.53} & 77.88 & \textbf{44.13} & \textbf{46.33} & 66.40 & \textbf{42.54} & \textbf{55.03} & \textbf{63.20} \\
\quad DeepSeek-V4-Flash    & 81.33 & \textbf{78.07} & 43.97 & 46.09 & \textbf{67.89} & 42.13 & 54.39 & 63.15 \\
\bottomrule
\end{tabular}
\end{table*}

Tab.~\ref{tab:sft-effectiveness} reveals two patterns. First, SFT training closes the gap between the compact executor and the frontier model: under a DeepSeek-V4-Flash backbone, Qwen3-1.7B-SFT improves over the naive Qwen3-1.7B by $+2.63$ EM and $+4.31$ LLM-judge, exceeding the DeepSeek-V4-Flash sub-agent. Second, this result generalizes: with GLM-5.1 as backbone, Qwen3-1.7B-SFT again matches the frontier sub-agent ($42.54$ vs.\ $42.13$ EM; $63.20$ vs.\ $63.15$ judge). Because a 1.7B-parameter model costs an order of magnitude less per token to serve than a frontier model, the SFT executor is the dominant choice across both backbones. Combined with the Pareto analysis of Sec.~\ref{sec:asymmetry-pareto}, these results confirm that a strong backbone paired with a compact, specialized executor advances the Pareto frontier for hierarchical search.
 
\subsection{Case Study: Decomposition Errors Propagate Through Correct Execution}
\label{sec:case-study}
 
The capacity sweeps above establish quantitatively that decomposition is the bottleneck. To understand \emph{why}, we qualitatively examine how a flawed decomposition propagates through otherwise correct execution.
 
\begin{figure}[t]
\centering
\includegraphics[width=\columnwidth]{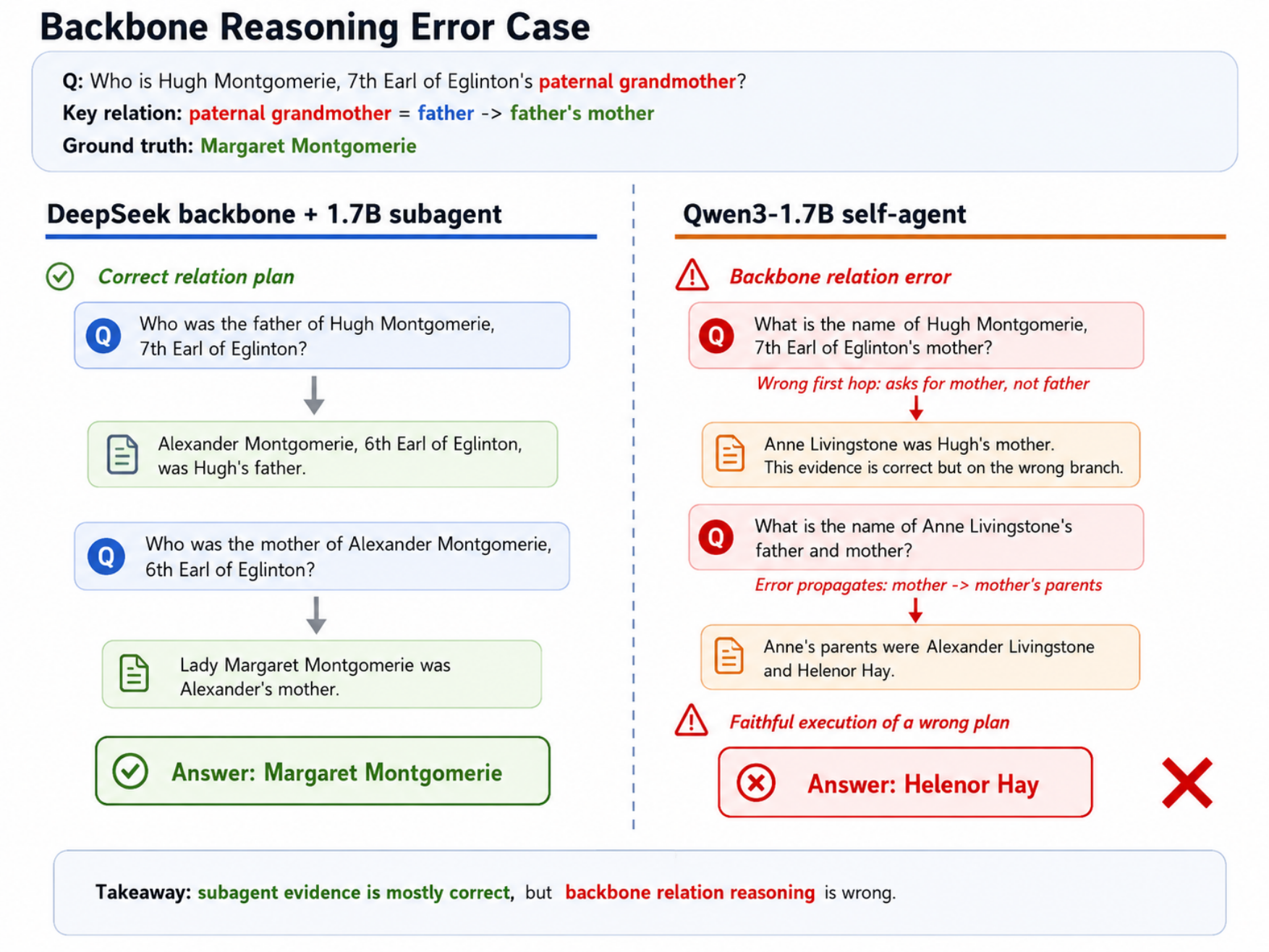}
\caption{Backbone reasoning error on a multi-hop question. Left: DeepSeek-V4-Flash correctly decomposes ``paternal grandmother'' into ``father $\rightarrow$ father's mother'', and the 1.7B sub-agent returns correct evidence at each hop. Right: Qwen3-1.7B misparses the relation, asking for the \emph{mother} instead of the \emph{father}; the sub-agent faithfully returns correct evidence for the wrong entity, and the error propagates to an incorrect final answer. The bottleneck is decomposition, not execution.}
\label{fig:case-study}
\end{figure}
 
Fig.~\ref{fig:case-study} presents a representative failure case. The question asks for Hugh Montgomerie, 7th Earl of Eglinton's \emph{paternal grandmother}, which requires decomposing the relation into two hops: first identify the father, then identify the father's mother. When DeepSeek-V4-Flash serves as backbone (left), it correctly parses this relational chain and issues two sequential sub-queries: ``Who was the father of Hugh Montgomerie?'' followed by ``Who was the mother of Alexander Montgomerie?'' The 1.7B sub-agent retrieves correct evidence at both hops, and the system arrives at the gold answer (Margaret Montgomerie).
 
When Qwen3-1.7B instead serves as the backbone in the same main--sub configuration (right), with the executor unchanged, it fails at the decomposition stage: it misparses ``paternal grandmother'' and asks for the subject's \emph{mother} rather than his \emph{father} at the first hop. Critically, the sub-agent's execution is not at fault. It faithfully retrieves a factually correct answer to the (wrong) question: Anne Livingstone was indeed Hugh's mother. But because the plan itself is flawed, the error propagates: the next query asks for Anne Livingstone's parents rather than for the father's mother, and the final answer (Helenor Hay) is incorrect despite every individual retrieval being accurate.
 
This case crystallizes the quantitative finding from the capacity sweeps. The sub-agent's retrieval and reading capability is sufficient even at 1.7B parameters; what fails is the backbone's relational reasoning during plan construction. A weak backbone produces a plausible but logically incorrect decomposition, and no amount of execution quality can recover from a fundamentally flawed plan. This explains why scaling the executor yields diminishing returns (Sec.~\ref{sec:execution-sweep}) while scaling the backbone produces large gains (Sec.~\ref{sec:delegation-sweep}): the bottleneck is the quality of the instructions, not the quality of their execution.

\section{Conclusion}
\label{sec:conclusion}
We studied the capacity allocation problem in multi-agent search by factorizing the task into delegation, execution, and answer generation, holding the last fixed as a confound control. Three findings emerge from controlled experiments across five multi-hop QA benchmarks.
 
First, multi-agent factorization itself is beneficial: separating delegation from execution improves EM from 4.5 to 8.6 points over a single-agent baseline, even when the same model fills both roles (Sec.~\ref{sec:multi-vs-single}). The gain is smallest for the frontier model, suggesting that role separation is most valuable when a single model lacks the capacity to jointly handle planning, retrieval, and synthesis.
 
Second, the two roles differ sharply in capacity sensitivity. Scaling the executor from 1.7B to frontier scale moves EM by only ${\sim}$2.6 points, whereas scaling the backbone over the same range moves EM by ${\sim}$11 points (Sec.~\ref{sec:execution-sweep}--\ref{sec:asymmetry-pareto}). Decomposition quality, not execution capability, governs the performance ceiling of a hierarchical search agent.
 
Third, targeted training advances the Pareto frontier: a 1.7B-parameter model trained via quality-filtered trajectory distillation matches a frontier sub-agent while consuming 37\% fewer sub-agent tokens, and generalizes across two different backbones (Sec.~\ref{sec:sft-executor}).
 
Our study has limitations. The analysis is conducted on English multi-hop QA with a fixed retrieval corpus; whether the asymmetry holds for broader search tasks (e.g., open-web research or multilingual settings) remains to be verified. The executor is also trained with supervised fine-tuning only; reinforcement learning over its interaction with the backbone may yield further gains.
 
We hope these results offer practitioners a concrete recipe for building multi-agent search systems, namely concentrating capacity at the delegation role while keeping execution compact, and provide a conceptual lens, role-aware capacity allocation, for future research on the division of labor in agentic systems.



\clearpage

\bibliography{iclr2025_conference}
\bibliographystyle{iclr2025_conference}

\clearpage
\appendix

\section{Choice of Answer-Generation Model}
\label{app:answerer-choice}

As discussed in Sec.~\ref{sec:experimental-setup}, we choose Qwen3-32B over DeepSeek as the fixed answer-generation model $\pi_A$. To validate this choice, we run both models on the 3{,}869 evaluation instances \emph{without} any retrieval, so that correct answers can only come from parametric memory. Tab.~\ref{tab:no-retrieval} reports the results.

\begin{table}[h]
\centering
\caption{No-retrieval accuracy on 3{,}869 evaluation instances. A high score indicates the model can answer from parametric memory alone, suggesting data contamination with the benchmarks.}
\label{tab:no-retrieval}
\begin{tabular}{lcc}
\hline
\textbf{Model} & \textbf{EM} & \textbf{F1} \\
\hline
DeepSeek without retrieval & 36.65 & 46.21 \\
Qwen3-32B without retrieval & 18.58 & 26.44 \\
\hline
\end{tabular}
\end{table}

DeepSeek achieves 36.65 EM without any retrieved evidence, indicating substantial overlap between its pretraining data and the evaluation benchmarks. Using it as $\pi_A$ would confound the experimental design: improvements from better delegation or execution could not be distinguished from the answerer already knowing the answer. Qwen3-32B, by contrast, reaches only 18.58 EM without retrieval, confirming that its correct answers depend on the evidence supplied by the search agent rather than on parametric memory. This makes it a suitable choice for a fixed answerer whose output faithfully reflects the quality of the collected evidence.

\section{LLM-as-Judge Prompt}
\label{app:judge-prompt}

\begin{figure}[htbp]
    \centering
    \includegraphics[width=0.8\linewidth]{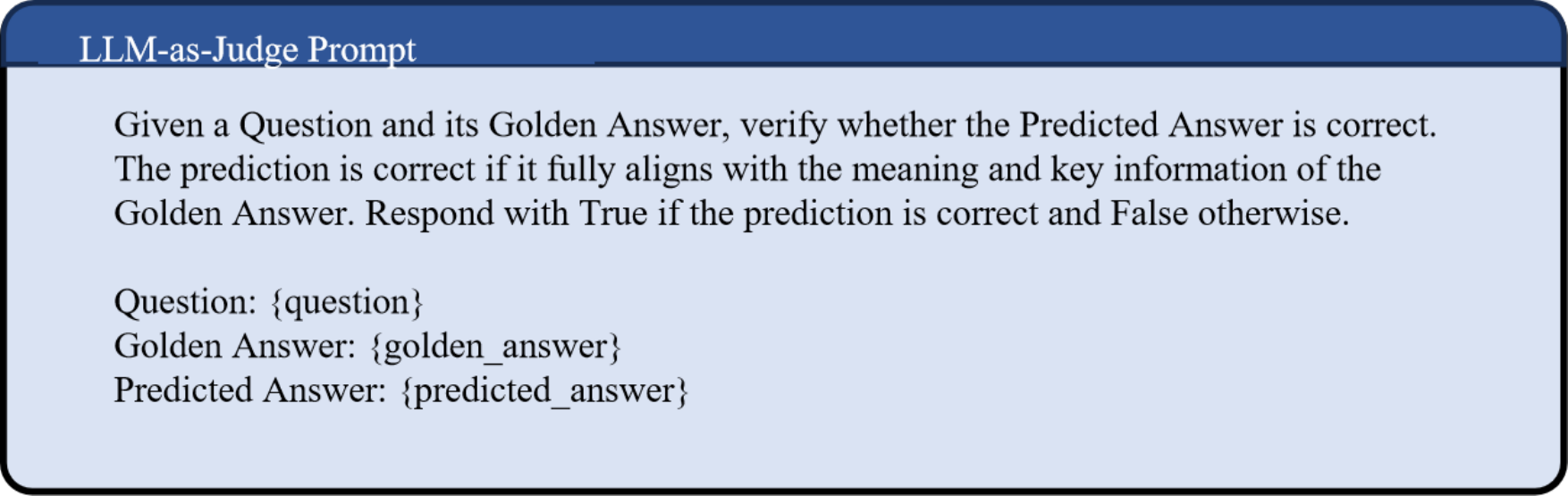}
    \caption{Prompt of LLM-as-Judge}
    \label{fig:prompt1}
\end{figure}

Following~\citet{song2025r1}, we use DeepSeek-V4-Flash as the judge model. The evaluation prompt is given in Fig. \ref{fig:prompt1}

\section{Role Prompts}
\label{app:role-prompts}

This appendix lists the system prompts of the delegation policy $\pi_D$ (backbone) and the execution policy $\pi_E$ (search sub-agent) described in Sec.~\ref{sec:role-factorization}. Both prompts are identical across all experimental conditions; only the model instantiating each role varies. When the backbone judges the collected evidence sufficient, it emits a stop signal rather than an answer; the delegation trajectory is then frozen, and the final answer is generated by the fixed answerer $\pi_A$ conditioned on the question, the dispatched sub-queries, and the returned reports (Sec.~\ref{sec:role-factorization}).

\subsection{Delegation Prompt (Backbone)}
\label{app:prompt-backbone}

\begin{figure}[h]
\centering
\includegraphics[width=\linewidth]{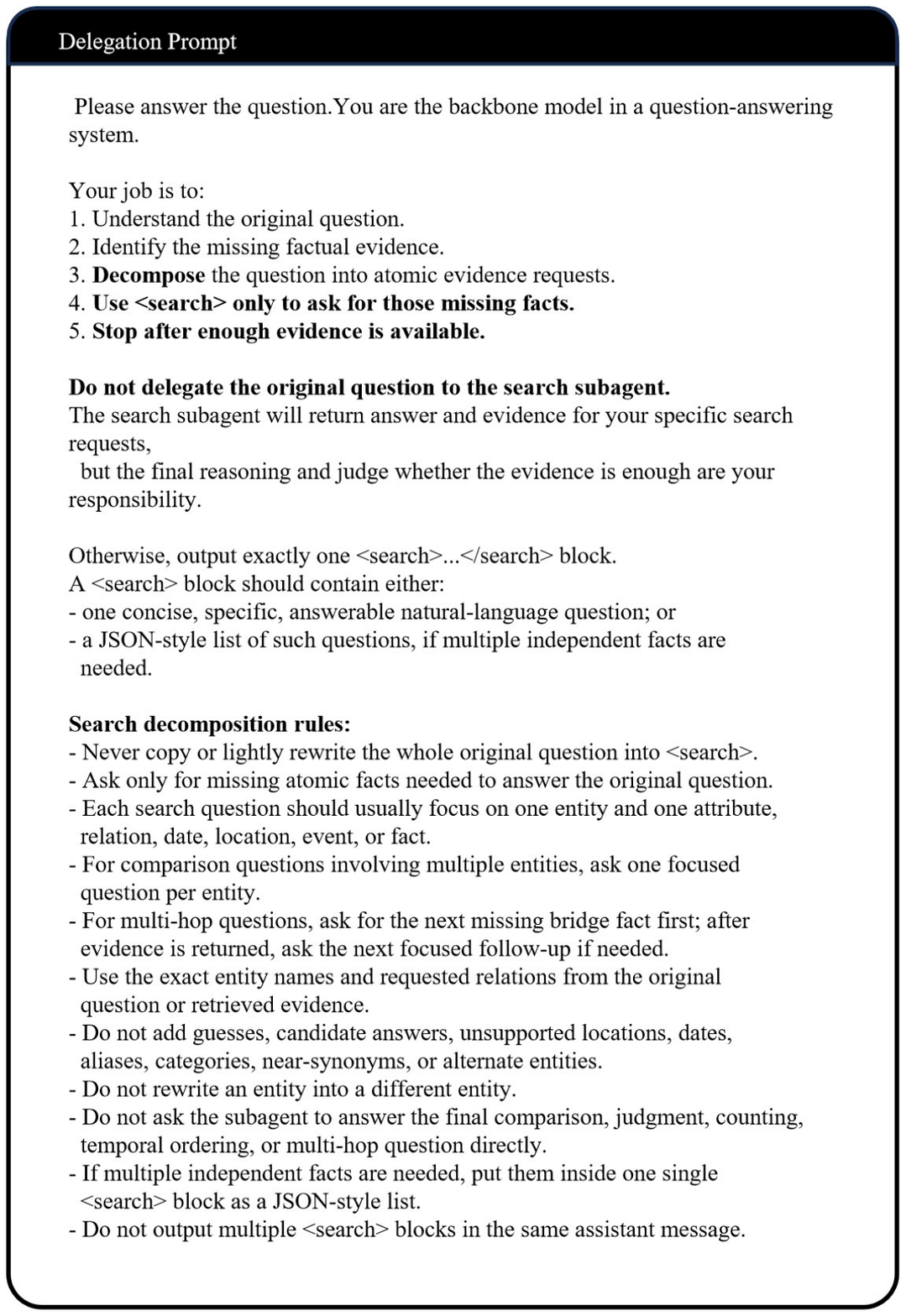}
\caption{Delegation prompt (part 1): role description, output format, and decomposition rules.}
\label{fig:delegation-prompt-1}
\end{figure}

\begin{figure}[h]
\centering
\includegraphics[width=\linewidth]{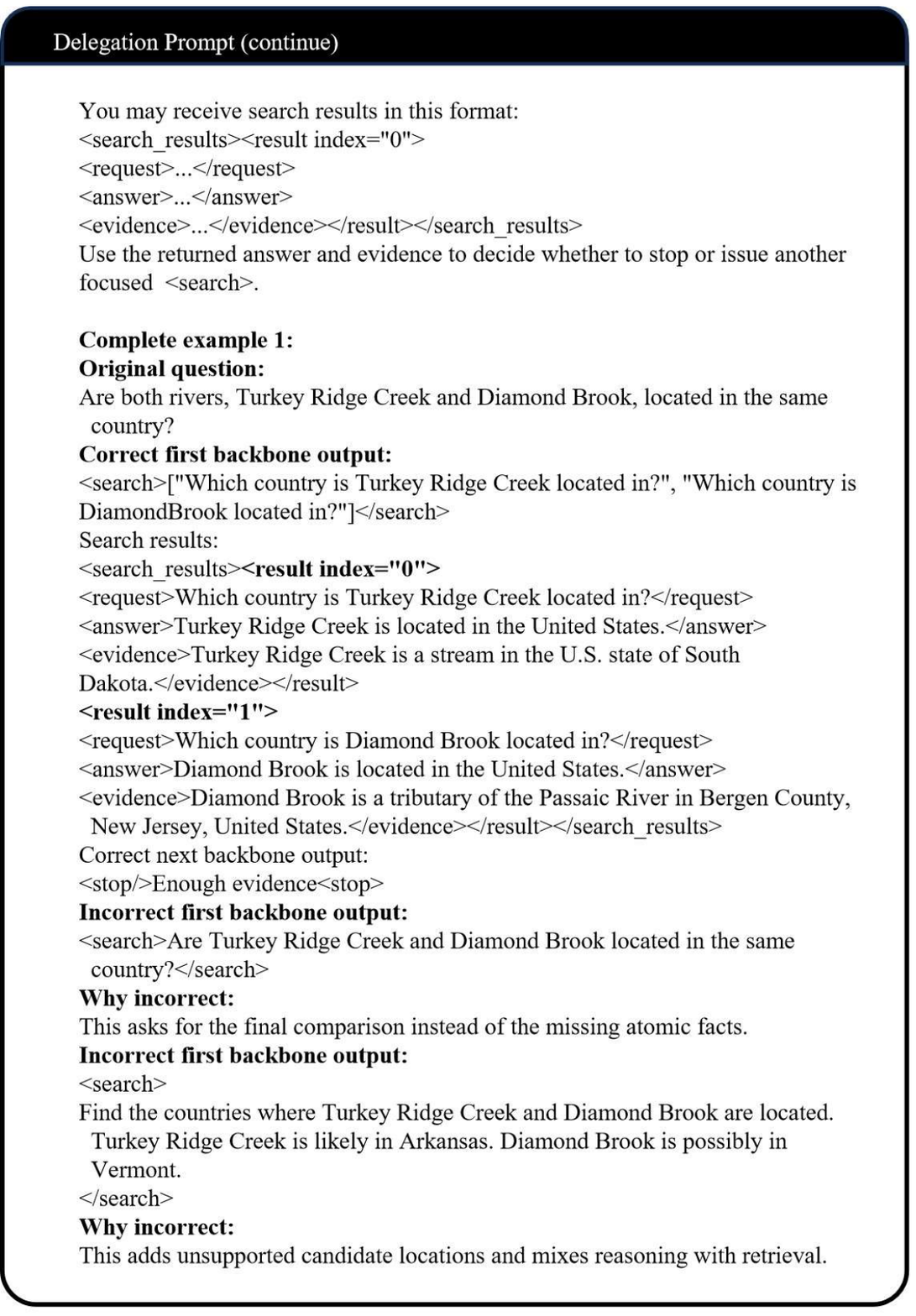}
\caption{Delegation prompt (part 2): worked example of a comparison question with parallel sub-queries.}
\label{fig:delegation-prompt-2}
\end{figure}

\begin{figure}[h]
\centering
\includegraphics[width=\linewidth]{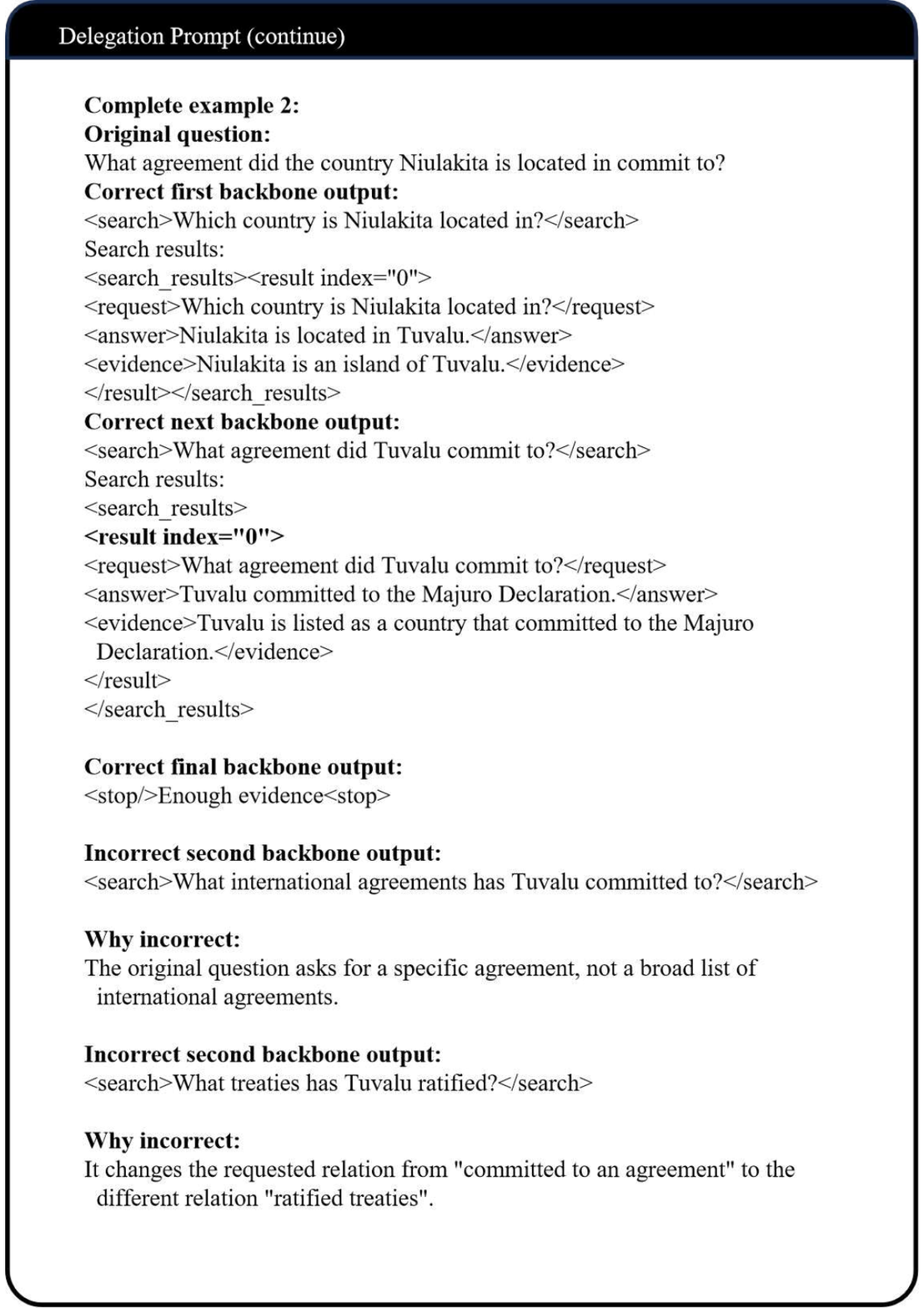}
\caption{Delegation prompt (part 3): worked example of a multi-hop question with sequential sub-queries.}
\label{fig:delegation-prompt-3}
\end{figure}

The full delegation prompt is shown in Figs.~\ref{fig:delegation-prompt-1}--\ref{fig:delegation-prompt-3}: Fig.~\ref{fig:delegation-prompt-1} gives the role description, output format, and decomposition rules; Figs.~\ref{fig:delegation-prompt-2} and \ref{fig:delegation-prompt-3} give two complete worked examples.

\subsection{Execution Prompt (Search Sub-Agent)}
\label{app:prompt-policy}

\begin{figure}[htbp]
\centering
\includegraphics[width=\linewidth]{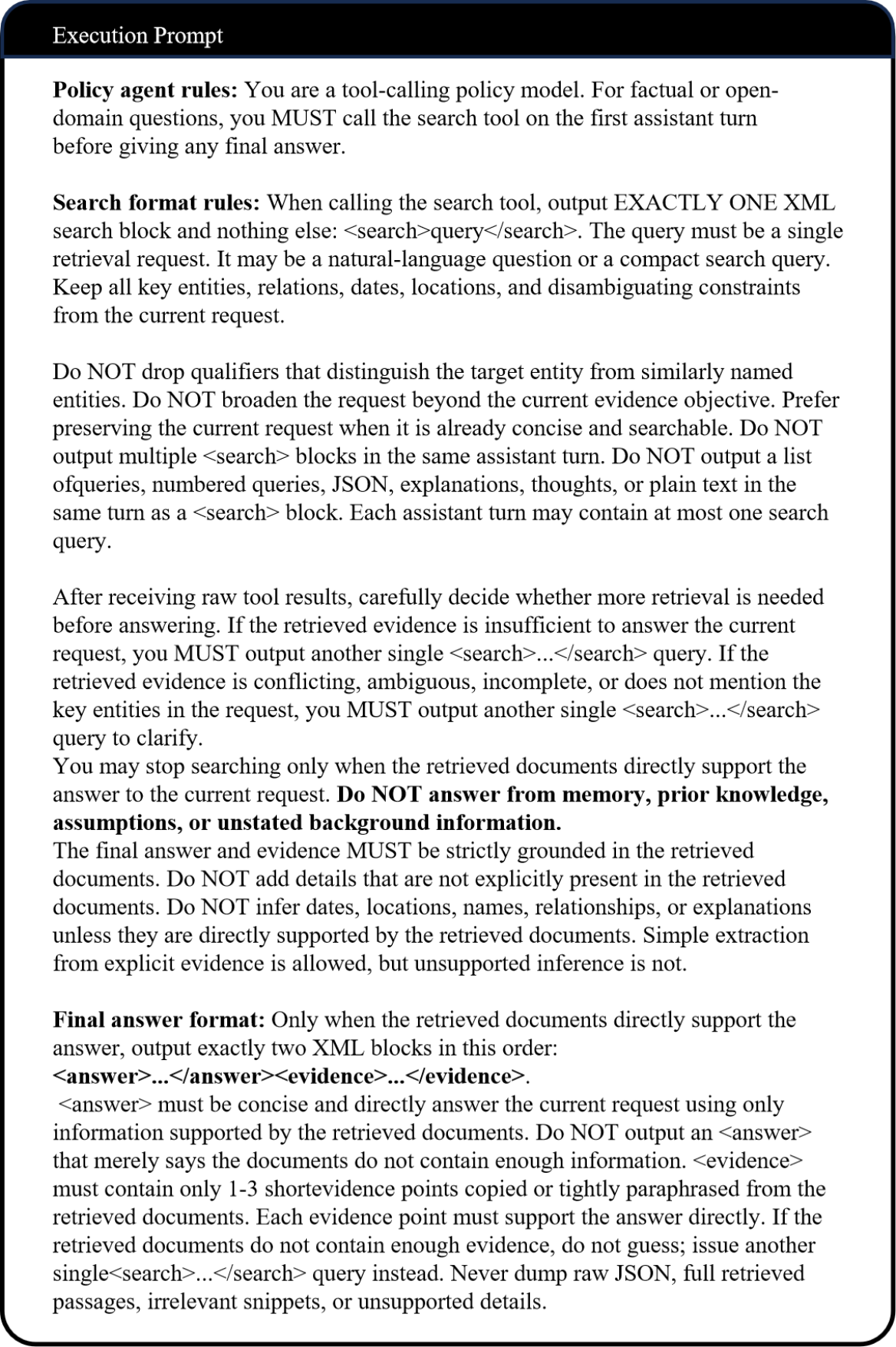}
\caption{Delegation prompt (part 3): worked example of a multi-hop question with sequential sub-queries.}
\label{fig:exe-prompt-3}
\end{figure}

The full execution prompt is shown in Fig.~\ref{fig:exe-prompt-3}

\section{SFT Data Composition}
\label{app:sft-data}

Rollouts are performed on 3{,}000 source questions drawn from the training splits of 2WikiMultihopQA, MuSiQue, and HotpotQA, disjoint from the evaluation instances. After the quality filtering of Sec.~\ref{sec:training-compact-executors}, the SFT corpus contains 2{,}168 training records derived from 1{,}591 unique source questions; a source question can contribute more than one record when trajectories from different strata are retained for it. Tab.~\ref{tab:sft-data} reports the composition by source dataset.

\begin{table}[h]
\centering
\caption{Composition of the filtered SFT corpus by source dataset (deduplicated by source question).}
\label{tab:sft-data}
\begin{tabular}{lcc}
\toprule
\textbf{Source dataset} & \textbf{SFT records} & \textbf{Unique source questions} \\
\midrule
2WikiMultihopQA & 910 & 627 \\
MuSiQue         & 615 & 489 \\
HotpotQA        & 643 & 475 \\
\midrule
Total           & 2{,}168 & 1{,}591 \\
\bottomrule
\end{tabular}
\end{table}

\section{Evaluation Set Construction}
\label{app:eval-cleaning}

From the 4{,}125 evaluation instances of the five benchmarks, we exclude 256 on which the delegation agent produces an answer without issuing any sub-query: such instances are solvable from parametric memory alone, and retaining them would leak knowledge that retrieval is meant to provide. Tab.~\ref{tab:eval-cleaning} reports the per-dataset breakdown; all experiments in Sec.~\ref{sec:analysis} use the remaining 3{,}869 instances.

\begin{table}[H]
\centering
\caption{Construction of the evaluation set: instances answerable by the delegation agent without any search are excluded.}
\label{tab:eval-cleaning}
\begin{tabular}{lccc}
\toprule
\textbf{Dataset} & \textbf{Original} & \textbf{Excluded} & \textbf{Remaining} \\
\midrule
2WikiMultihopQA & 1{,}000 & 9   & 991 \\
HotpotQA        & 1{,}000 & 37  & 963 \\
MuSiQue         & 1{,}000 & 12  & 988 \\
PopQA           & 1{,}000 & 182 & 818 \\
Bamboogle       & 125     & 16  & 109 \\
\midrule
Total           & 4{,}125 & 256 & 3{,}869 \\
\bottomrule
\end{tabular}
\end{table}

\end{document}

%% file: math_commands.tex
\usepackage{amsmath,amsfonts,bm}

\def\eqref#1{equation~\ref{#1}}

\def\1{\bm{1}}

\DeclareMathAlphabet{\mathsfit}{\encodingdefault}{\sfdefault}{m}{sl}
\SetMathAlphabet{\mathsfit}{bold}{\encodingdefault}{\sfdefault}{bx}{n}

